\documentclass[10pt,twocolumn,letterpaper]{article}

\usepackage[pagenumbers]{cvpr} 

\usepackage{graphicx}
\usepackage{amsmath}
\usepackage{amssymb}
\usepackage{color}
\usepackage{xcolor}
\usepackage{booktabs}
\usepackage{amsfonts}
\usepackage{bbm}
\usepackage{multirow}
\usepackage{pifont}
\usepackage[accsupp]{axessibility}
\usepackage[titletoc]{appendix}
\newcommand{\cmark}{\ding{51}}%
\newcommand{\xmark}{\ding{55}}%

\definecolor{blue-violet}{rgb}{0.54, 0.17, 0.89}
\definecolor{red-violet}{rgb}{0.89, 0.17, 0.17}

\newcommand{\gao}[1]{{#1}}

\usepackage[pagebackref,breaklinks,colorlinks]{hyperref}

\usepackage[capitalize]{cleveref}
\crefname{section}{Sec.}{Secs.}
\Crefname{section}{Section}{Sections}
\Crefname{table}{Table}{Tables}
\crefname{table}{Tab.}{Tabs.}

\def\cvprPaperID{5116} 
\def\confName{CVPR}
\def\confYear{2023}

\begin{document}

\title{SurfelNeRF: Neural Surfel Radiance Fields for Online Photorealistic Reconstruction of Indoor Scenes}

\author{Yiming Gao \qquad Yan-Pei Cao \qquad Ying Shan \\
ARC Lab, Tencent PCG \\
{\tt\small gaoym9@mail3.sysu.edu.cn\qquad caoyanpei@gmail.com\qquad yingsshan@tencent.com}
}
\maketitle
\begin{abstract}
Online reconstructing and rendering of large-scale indoor scenes is a long-standing challenge. SLAM-based methods can  reconstruct 3D scene geometry progressively in real time but can not render photorealistic results.  While NeRF-based methods produce promising novel view synthesis results, their long offline optimization time and lack of geometric constraints pose challenges to efficiently handling online input. Inspired by the complementary advantages of classical 3D reconstruction and NeRF, we thus investigate marrying explicit geometric representation with NeRF rendering to achieve efficient online reconstruction and high-quality rendering. We introduce SurfelNeRF, a variant of neural radiance field which employs a flexible and scalable neural surfel representation to store geometric attributes and extracted appearance features from input images. We further extend the conventional surfel-based fusion scheme to progressively integrate incoming input frames into the reconstructed global neural scene representation. In addition, we propose a highly-efficient differentiable rasterization scheme for rendering neural surfel radiance fields, which helps SurfelNeRF achieve $10\times$ speedups in training and inference time, respectively. Experimental results show that our method achieves the state-of-the-art 23.82 PSNR and 29.58 PSNR on ScanNet in feedforward inference and per-scene optimization settings, respectively.\footnote{Project website: \url{https://gymat.github.io/SurfelNeRF-web} }

\end{abstract}

\section{Introduction}
\label{sec:intro}

\begin{figure}[t]
	\centering
	\begin{minipage}[t]{1\linewidth}
		\centering
		\includegraphics[width=1\textwidth]{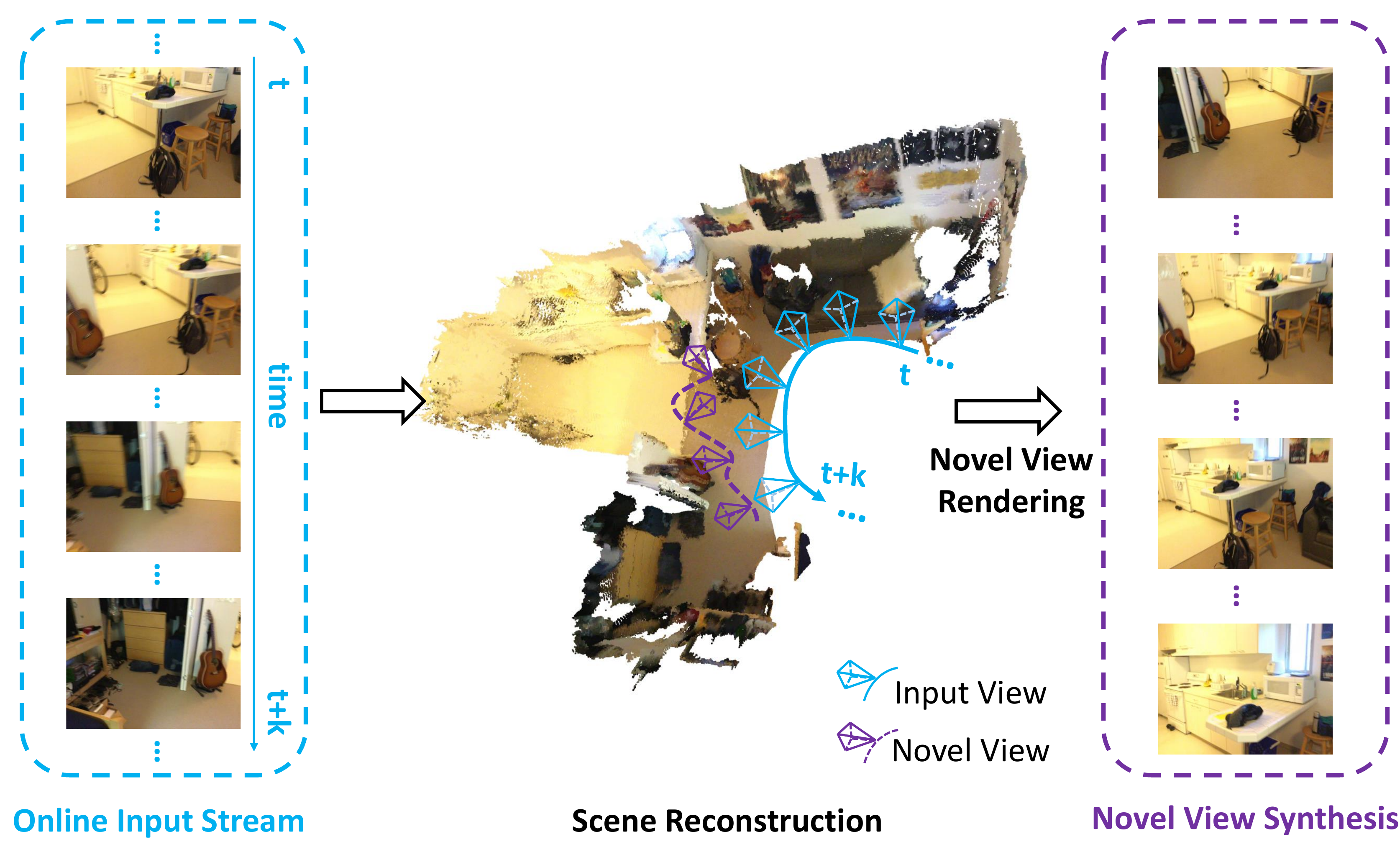}
	\end{minipage}
	\caption{ Examples to illustrate the task of online photorealistic reconstruction of an indoor scene. The \textit{online photorealistic} reconstruction of large-scale indoor scenes: given an online input image stream of a previously unseen scene, the goal is to progressively build and update a scene representation that allows for high-quality rendering from novel views.
	}
	\label{fig:teaser}
\end{figure}

\begin{table*}[t]
	\small
	\center
	\setlength{\tabcolsep}{5 pt}
	\begin{center}
		\begin{tabular}{c|ccccccc}
			\toprule
			Methods      & Representation  & Generalization & Real-time Rendering & Online Fusion &  Scalability \\ \hline
			DeepSurfel~\cite{mihajlovic2021deepsurfels}  & Surfels		  & \xmark        & \cmark         & \xmark &  \xmark \\
			Instant-NGP~\cite{mueller2022instant} & Hash Grids       & \xmark        & \cmark         & \xmark & \xmark \\
			PointNeRF~\cite{xu2022point}    & Point Clouds    & \cmark     & \xmark      & \xmark & \cmark \\
			VBA~\cite{clark2022volumetric}    & B+ Trees     & \xmark     & \xmark     & \cmark    & \cmark \\
			NeRFusion~\cite{zhang2022nerfusion}    & Voxel Grids     & \cmark     & \xmark         & \cmark & \xmark \\
			Ours         & Surfels		  & \cmark     & \cmark         & \cmark &  \cmark \\
			\bottomrule
		\end{tabular}
	\end{center}
	\caption{Comparison of representation and features with existing methods.}
 \vspace{-4 mm}
	\label{tab:comparison_features}
\end{table*}

Large-scale scene reconstruction and rendering is a crucial but challenging task in computer vision and graphics with many applications. Classical visual simultaneous localization and mapping (SLAM) systems~\cite{KinectFusion,whelan2015elasticfusion,weder2020routedfusion,BundleFusion,Cao_TOG_2018,10.1145/3503926} can perform real-time 3D scene reconstruction. However, they usually represent the scene geometry as solid surfaces and appearance as vertex color or texture maps; thus, the reconstructed results fail to fully capture the scene content and cannot be used for photorealistic rendering. Recently, neural radiance fields (NeRF) and its variants~\cite{mildenhall2020nerf, barron2021mip, tancik2022block,xiangli2021citynerf, barron2022mip,rematas2021urban} have achieved unprecedented novel view synthesis quality on both object-centric and large-scale scenes. However, NeRFs suffer from  long per-scene optimization time and slow rendering speed, especially for large-scale scenes. Although recent advances~\cite{mueller2022instant,sun2021direct,ReluField_sigg_22,chen2022tensorf,liu2022devrf} achieve faster optimization and rendering via incorporating explicit representations, they still require gathering all input images in an \textit{offline} fashion before optimizing each scene.

In this paper, we target the challenging task of \textit{online photorealistic} reconstruction of large-scale indoor scenes: given an online input image stream of a previously unseen scene, the goal is to progressively build and update a scene representation that allows for high-quality rendering from novel views. The online setting can unlock a variety of real-time interactive applications, providing crucial immediate feedback to users during 3D capture. 
However, this task brings multiple extra requirements, including the scalability of the underlying scene representation, the ability to perform on-the-fly updates to the scene representation, and optimizing and rendering at interactive framerates. Recently, NeRFusion~\cite{zhang2022nerfusion} followed NeuralRecon~\cite{NeuralRecon} to unproject input images into local sparse feature volumes, fusing them to a global volume via Gated Recurrent Units (GRUs), and then generating photorealistic results from the global feature volume via volume rendering. However, updating the sparse volumetric feature involves computationally heavy operations; the volume rendering is also very slow since it requires \textit{hundreds} of MLP evaluations to render a pixel. Thus, although NeRFusion achieves efficient online scene reconstruction, it still needs dozens of seconds to render a frame. VBA~\cite{clark2022volumetric} is another recent approach to online photorealistic scene capture, but it only applies to object-centric scenes. We compare with representation and key features used in online photorealistic rendering with existing methods, which is shown in Tab.~\ref{tab:comparison_features}.

We propose surfel-based neural radiance fields, SurfelNeRF, for online photorealistic reconstruction and rendering of large-scale indoor scenes. Surfels (\textbf{surf}ace \textbf{el}ements)~\cite{pfister2000surfels} are point primitives containing geometric attributes, e.g., position, color, normal, and radius. We extend this representation to \textit{neural surfels}, storing extra neural features that encode the neural radiance field of the target scene. Compared with volumetric representations, neural surfels are more compact and flexible and can easily scale to large scenes. Besides, we further employ a fast and differentiable rasterization process to render neural surfel radiance fields, which produces a pixel with only \textit{a few} MLP evaluations based on the rasterized neural surfels. 

Inspired by classical real-time surfel-based geometric reconstruction methods~\cite{whelan2015elasticfusion,keller2013real,Cao_TOG_2018}, we propose an efficient neural radiance field fusion method to progressively build the scene representation by integrating neighboring neural surfels. Unlike point-based representations~\cite{xu2022point,npbg,Rakhimov_2022_CVPR,ruckert2021adop} that are computationally heavy when finding neighboring points, it is easier to locate overlapping surfels and then merge neural features from multiview observations. By coupling the SurfelNeRF representation, the efficient neural surfel fusion approach, and the fast neural surfel rasterization algorithm, we achieve high-quality, photorealistic 3D scene reconstruction in an online manner.

We conduct experiments on the large-scale indoor scene dataset ScanNet~\cite{DBLP:conf/cvpr/DaiCSHFN17}, which contains complex scene structures and a large variety of scene appearances. We train the SurfelNeRF end-to-end across the scenes on the ScanNet, obtaining a generalizable model that enables both feedforward inference on unseen data and per-scene fine-tuning. We demonstrate in experiments that the proposed SurfelNeRF achieves favorably better rendering quality than the state-of-the-art approaches in both feedforward and fine-tuning settings while maintaining high training and rendering efficiency. We believe the proposed online photorealistic reconstruction framework has great potential in practical applications.

\section{Related Work}

\noindent\textbf{Novel view synthesis.}
Novel view synthesis is a challenging task for computer vision and computer graphics. Previous methods utilize geometry-based view synthesis. 
Zhou~\etal~\cite{zhou2018stereo} proposes to use the multiplane image (MPI) representation for novel view synthesis. 
Recently, neural radiance fields~\cite{mildenhall2020nerf} and neural-based geometry representation have attracted more interest due to high-quality rendering. 
Recent methods propose to associate neural features on geometry structure and then synthesize novel views via rendering rasterized features, such as point cloud~\cite{npbg,ruckert2021adop}, texture~\cite{thies2019deferred} and surfel~\cite{mihajlovic2021deepsurfels}. 
\gao{
DeepSurfel performs online appearance fusion only on a fixed given geometry during online fusion, while our SurfelNeRF performs online \textit{appearance and geometry} integration. To render outputs, DeepSurfel rasterizes surfels and then passes them through 2D CNN filters, while our SurfelNeRF performs neural volume rendering which could achieve higher-quality rendering. Some recent works~\cite{chen2022mobilenerf,hedman2021baking,li2021neulf,yu2021plenoctrees} focus on real-time neural rendering methods.
}
However, these methods all require per-scene optimization, which makes them impractical to handle novel scenes. In contrast, our SurfelNeRF is generalizable, which can reconstruct new scenes via direct feedforward inference.

\noindent\textbf{Neural Radiance Fields.}
Recently, neural radiance fields (NeRF) methods~\cite{mildenhall2020nerf, barron2021mip, xiangli2021citynerf} achieve promising results on novel view synthesis, but it requires expensive per-scene optimization. Some methods proposed to reduce the long per-scene optimization time via deferred neural rendering~\cite{hedman2021baking,zhang2022digging}, hash table~\cite{mueller2022instant} or geometry prior~\cite{liu2020neural}. Recent, generalization NeRF methods~\cite{wang2021ibrnet, chen2021mvsnerf, yu2021pixelnerf, xu2022point, cao2022fwd} are proposed to produce novel view without per-scene optimization by learning a radiance fields representation from offline given images of novel scenes. PixelNeRF~\cite{yu2021pixelnerf} and IBRNet~\cite{wang2021ibrnet} render novel view images using
 volume rendering based on corresponding warped features from nearby given images. MVSNeRF~\cite{chen2021mvsnerf} reconstructs local neural volumes from local nearby views, and then conducts volume rendering based on volumes. FWD~\cite{cao2022fwd} renders novel views by warping all input images to the target view, and fusing and refining features of these warped inputs.
 However, these methods are generalizable on local small-scale scenes, and require taking all input images before rendering novel views in an offline fashion.
 PointNeRF~\cite{xu2022point} learns to reconstruct a point-based neural radiance field for the novel scene. However, point-based representation is redundant, especially on large-scale scenes, and the point clouds cannot be updated based on multi-view input. Unlike these works, our method can represent not only large-scale scenes but also update compact surfel representation in an online fashion.

\noindent\textbf{Online scene reconstruction and rendering.}
Most online scene reconstruction methods~\cite{NeuralRecon, bozic2021transformerfusion, whelan2015elasticfusion, Zhu2022CVPR,zou2022mononeuralfusion} focus on 3D reconstruction only via TSDF-based fusion~\cite{NeuralRecon, bozic2021transformerfusion}, surfel-based fusion~\cite{whelan2015elasticfusion} and SLAM-based reconstruction~\cite{Zhu2022CVPR}. However, they cannot render photorealistic novel view synthesis. Recently, some methods~\cite{zhang2022nerfusion,sajjadi2022scene,clark2022volumetric} employ radiance fields to tackle online scene representation for high-quality rendering. VBA~\cite{clark2022volumetric} builds a neural dynamic B+Tree to represent scenes and apply volume rendering on it, but it still requires per-scene optimization. SRT~\cite{sajjadi2022scene} represents scenes with set-latent with a transformer encoder and decoder, but it cannot synthesize high-resolution images, which makes it not ready for real-world applications. NeRFusion~\cite{zhang2022nerfusion} represents large-scale scenes with neural voxel grids, update them via GRU in an online fashion, and renders results via slow volume rendering. However, grid-based representation is wasteful for large-scale scenes, and is inefficient during features update due to features in empty areas are also updated. In contrast, our surfel representation is compact for large-scale scenes, and surfel-based fusion and corresponding rendering are more efficient.

\begin{figure*}[t]
	\centering
	\begin{minipage}[t]{0.9\linewidth}
		\centering
		\includegraphics[width=1\textwidth]{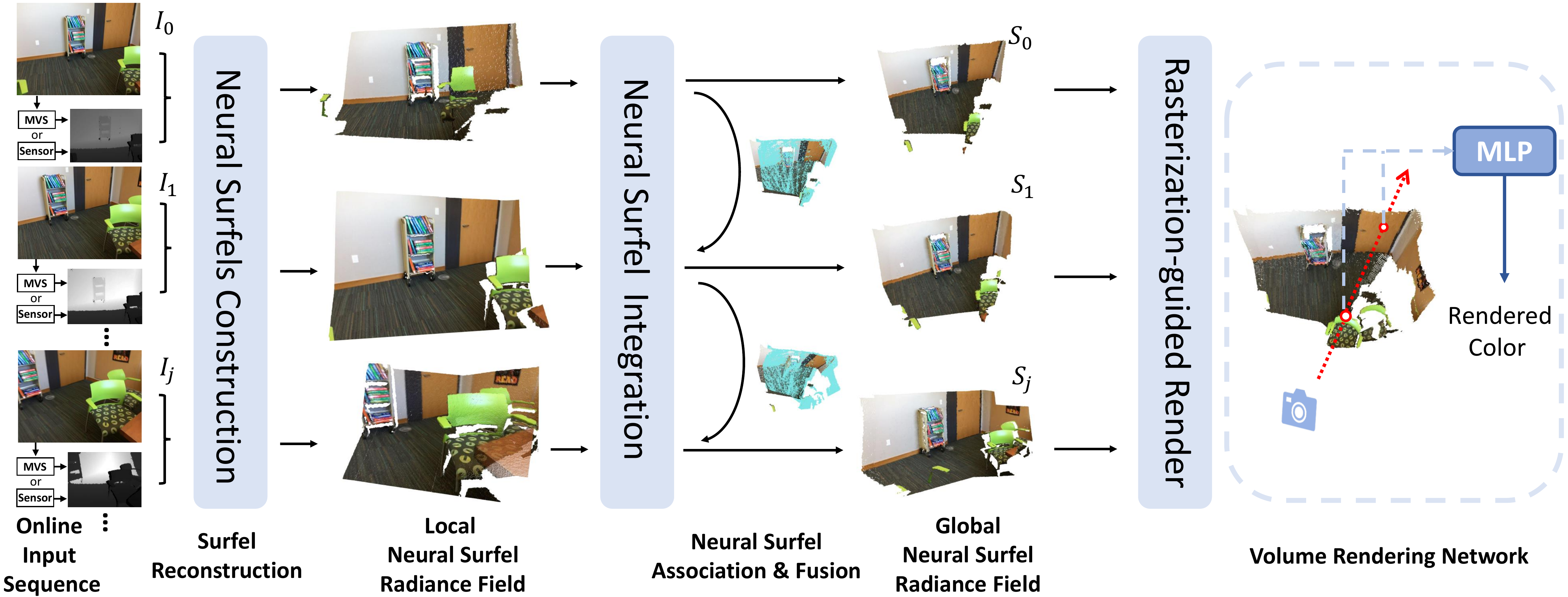}
	\end{minipage}
	\caption{Overview of our SurfelNeRF. Given an online input stream of image sequences, we first reconstruct a surfel representation associated with neural features to build a local neural surfel radiance field for each input keyframe. Then the neural surfel radiance field integration is used to fuse the input local neural surfel radiance field into the global neural surfel radiance field by updating both surfel position and features. More specifically, input local neural surfels associated with global surfels are fused to global surfels with corresponding global ones, and remaining local surfels without corresponding ones are added to the global model. Furthermore, the novel views can be rendered from updated global surfels via our efficient rasterization-guided render. Our proposed rasterization-guided renderer renders color only on the intersection points of ray and surfels, which is faster than volume rendering.
	}
	\label{fig:overview}
\end{figure*}

\section{Method}

Unlike most existing methods~\cite{mildenhall2020nerf,mueller2022instant,xu2022point} that build neural radiance fields in an offline fashion, we aim for the task of online photorealistic scene reconstruction. Given a sequence of images $\{I_n\}$, the goal is to progressively build the neural scene representation on the fly as the input images arrive.

To achieve this goal, we present SurfelNeRF, an online neural reconstruction framework based on the  neural surfel radiance fields representation. The insight of our method is that the classical explicit surfel representation is efficient for both reconstruction and rendering, has compact memory footprints, has the scalability to large and complex scenes, and can be naturally extended to equip neural attributes. Meanwhile, one can integrate multi-frame information and perform incremental updates on neural surfels in a \textit{geometrically meaningful} way, and the operations involved can be streamlined.

Figure~\ref{fig:overview} gives an overview of our proposed method. The framework first learns to generate per-view local neural surfel radiance fields $\mathcal{S}_t$ for each input frame $I_t$. We then adopt the conventional surfel fusion process~\cite{whelan2015elasticfusion,keller2013real,Cao_TOG_2018} to the neural setting, parameterizing the neural surfel fusion procedural as a neural network $N_f$. Through $N_f$, per-view representations $\{\mathcal{S}_t\}$ can be integrated into a global neural surfel radiance field $\mathcal{S}_g$ on the fly, aggregating geometric and neural attributes. The neural surfel radiance fields can be efficiently rendered to photorealistic images using a differentiable rasterizer without extra optimization. The framework can be trained end-to-end on large-scale indoor scene datasets. Once trained, it can generalize to previously unseen data and perform online photorealistic reconstruction; if needed, the rendering quality can be further improved by a short period time of fine-tuning.

\subsection{Neural Surfel Radiance Field Construction}
\label{sec:nerual_surfel_construction}

\noindent\textbf{Surfel Representation.} Adopted from classical 3D reconstruction research, each surfel $s^n$  is equipped with the following geometric attributes: a position $\mathbf{p}^n \in \mathbb{R}^3$, normal $\mathbf{n}^n \in \mathbb{R}^3$, weight $w^n\in \mathbb{R}$ and radius $r^n \in \mathbb{R}$. The weight $w^n$ measures the confidence of a surfel based on the distance of the current depth measurement from the camera center. The radius $r^n$ of each surfel is designed to represent the local support area around the projected 3D position while minimizing visible holes. The weight and radius of each surfel are initialized following~\cite{whelan2015elasticfusion}.
Besides geometric properties, we further associate each surfel with neural features $\mathbf{f}^n \in \mathbb{R}^N$ from input images for modeling semi-transparent structures and view-dependent effects. Although a similar representation has been explored in DeepSurfels~\cite{mihajlovic2021deepsurfels}, we construct neural surfels in a feedforward manner and incorporate them in progressive scene reconstruction.

\noindent\textbf{Local Neural Surfel Radiance Field.}
Given an input frame $I_t$, camera parameters $\mathbf{c}_t$, and estimated depth $D_t$, the corresponding \textit{local} neural surfel radiance field $\mathcal{S}_t$ can be constructed by unprojecting pixels in the input image to 3D space using camera parameters, then estimating corresponding normal, weight, radius, and neural features.  
The depth $D_t$ can be captured from sensors or estimated via off-the-shelf MVS-based methods~\cite{yao2018mvsnet}, which might be incomplete and noisy. Thus, we employ a U-Net structure~\cite{ronneberger2015u} to refine the depth following~\cite{cao2022fwd}.

\noindent\textbf{Neural Features on Surfels.} We estimate neural features $\mathbf{f}^n$ of surfels by extracting corresponding pixel features in the input image. We employ a 2D CNN to extract features of input images. Specifically, we use a MasNet following~\cite{zhang2022nerfusion, NeuralRecon} to extract multi-scale view-dependent features.

\subsection{Online Integration of Neural Surfels}
\label{sec:surfel_fusion}

Based on the neural surfel representation, we adopt a geometry-guided fusion scheme to efficiently and progressively integrate local neural surfel radiance fields into a global model. Compared with previous volumetric fusion methods~\cite{clark2022volumetric, zhang2022nerfusion}, our surfel-based solution offers the advantages of 1) efficiency - the fusion operation only needs to be performed at surfel locations and 2) scalability - the memory footprint only grows linearly to the number of neural surfels.

In particular, our fusion scheme can be divided into two steps. Firstly, for each local neural surfel, we determine whether it should be added as a  global neural surfel or should be merged with corresponding global surfels. Secondly, we insert it into the global model or perform updates to corresponding global surfels with local features.

\noindent\textbf{Neural Surfel Association.} 
Given an input image frame $I_t$ and corresponding camera parameter, each surfel in global surfels can be projected as an ellipse onto the image plane of the current input camera view. For each pixel in the input image, we obtain top-$K$ global surfels that are overlapped with it after rasterization. We discard irrelevant surfels with a normal filter. Then we determine the incremental surfel by comparing the depth distance between respective top-M global surfels from near to far step by step. If the minimal distance between a local surfel and its global surfel candidates is smaller than a threshold $\delta_{depth}$, then we determine that this input surfel needs to be merged; otherwise, it should be directly added to global surfels. The overall workflow follows~\cite{whelan2015elasticfusion}, and $K$ is set to 8.

\noindent\textbf{Neural Surfel Fusion.}
After the surfel association, there are a set of input surfels that need to be added directly and another set of surfels that will be merged with corresponding global surfels. For the former set of surfels, we directly construct neural surfels (Sec.~\ref{sec:nerual_surfel_construction}) and then add them to global surfels. 
For the latter surfels, we use a GRU network to update neural features between the features $\mathbf{f}_t^{\mathtt{merge}}$ of input surfels and features $\mathbf{f}_{t-1}^{\mathtt{corrs}}$ of corresponding global surfels, which is given as
\begin{equation}
	\mathbf{f}_{t}^{\mathtt{corrs}} = \mathtt{GRU}(\mathbf{f}_t^{\mathtt{merge}}, \mathbf{f}_{t-1}^{\mathtt{corrs}}),
	\label{eq:gru}
\end{equation}
where $\mathbf{f}_t^{\mathtt{merge}}$ and $\mathbf{f}_{t-1}^{\mathtt{corrs}}$ are input and hidden state of GRU respectively. We then update the geometric attributes of these global surfels via a weighted sum. The update rule of each respective pair of local surfel $s^i_t$ and global surfel $s^j_{t-1}$ can be expressed by
\begin{equation}
	\begin{split}
		\mathbf{p}^j_t &= \frac{w_t^i\mathbf{p}^i_t + w_{t-1}^{j}\mathbf{p}^j_{t-1}}{w_t^i + w_{t-1}^{j}}, \\
		\mathbf{n}^j_t &= \frac{w_t^i\mathbf{n}^i_t + w_{t-1}^{j}\mathbf{n}^j_{t-1}}{w_t^i + w_{t-1}^{j}}, \\
		r^j_t &= \frac{w_t^ir^i_t + w_{t-1}^{j}r^j_{t-1}}{w_t^i + w_{t-1}^{j}}, \\
		w^j_t &= {w_t^i + w_{t-1}^{j}}, 
	\end{split}
	\label{eq:surfel_update}
\end{equation}
where $\mathbf{p}^j_t$, $\mathbf{n}^j_t$, $r^j_t$ and $w^j_t$ are the updated position, normal, radius, and weight of the global surfels, respectively.

\subsection{Rendering Neural Surfel Radiance Fields}

\begin{figure}[t]
	\centering
	\begin{minipage}[t]{1\linewidth}
		\centering
		\includegraphics[width=1\textwidth]{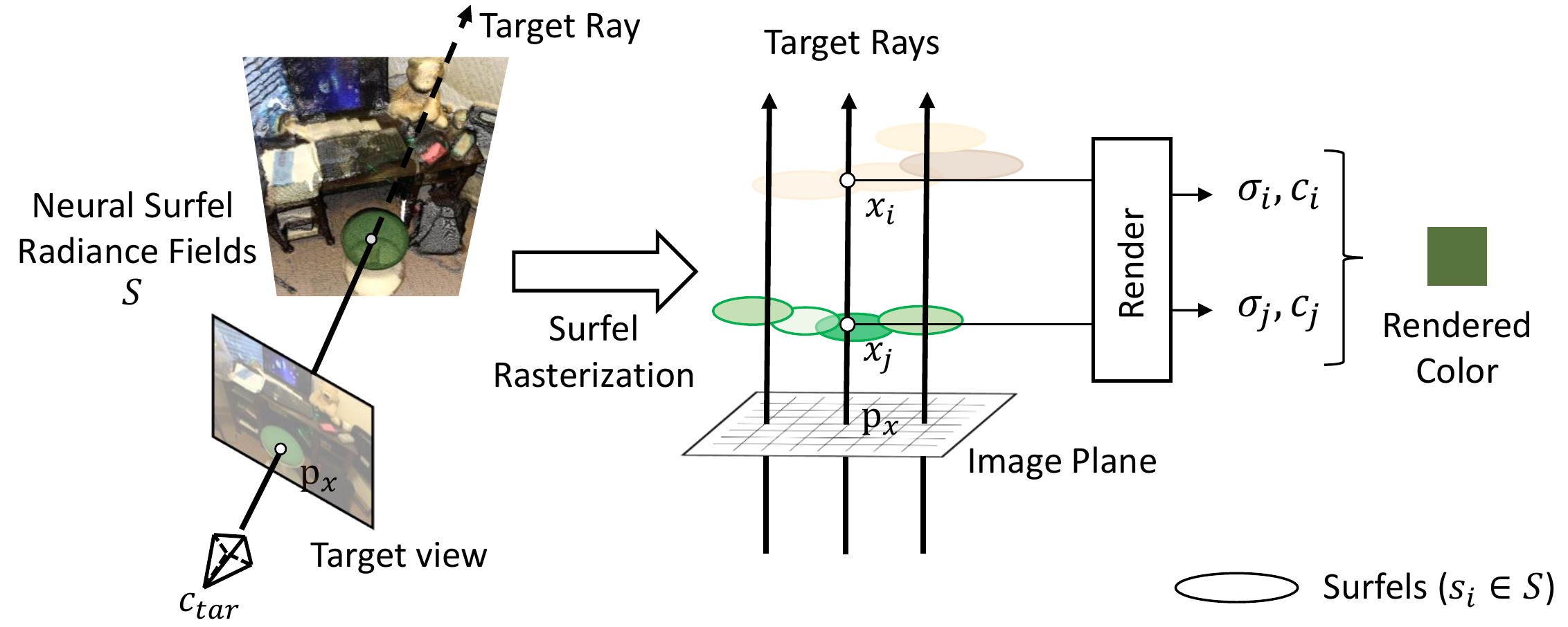}
	\end{minipage}
	\caption{Our efficient rasterization-guided surfel rendering for neural surfel radiance fields is visualized using a toy model as an example. We begin by rasterizing neural surfels into the image plane, and then render the ray using shading points on the intersections between the ray and corresponding surfels. This method uses a combination of rasterization and ray tracing to render view-dependency results with surfel representation and 3D rendering.
	}
	\label{fig:surfel_examples}
\end{figure}


We propose a fast rasterization-based rendering algorithm for neural surfel radiance fields (see Figure.~\ref{fig:surfel_examples}). Given a neural surfel radiance field $\mathcal{S}$ and the camera parameter $\mathbf{c}_{tar}$ of a target view, we first conduct a differentiable rasterization process on the neural surfels to project them on the target image plane. For a pixel $p_x$ in the target image, we gather the neural surfels $\{s_i\}_{i=1}^M$ that cover it and render its color $C(p_x)$ based on a few shading points $\{\mathbf{x}_i\}_{i=1}^M$ that are the intersections between $\{s_i\}_{i=1}^M$ and pixel ray, using the following volume rendering formula~\cite{mildenhall2020nerf}:
\begin{equation}
	\begin{split}
		C(p_x) &= \sum^M\tau_i(1 - \exp(-\sigma_i\Delta_i))c_i, \\
		\tau_i &= \exp(-\sum_{j=1}^{i-1}\sigma_j\Delta_j), 
	\end{split}
	\label{eq:volume_render}
\end{equation}
where $\tau$ indicates the accumulated transmittance; $\sigma_i$ and $c_i$  indicate the volume density and radiance on shading point $\mathbf{x}_i$; $\Delta_j$  is the distance between adjacent shading points.
For each shading point $x_i$, we regress its volume density $\sigma_i$ and radiance $c_i$ based on its view direction $\mathbf{d}$ and shading point features $\mathbf{f}^{i}(\mathbf{x}_i)$ associated from surfel $s_i$, which can be formulated as:
\begin{equation}
    \sigma_i, c_i = \mathtt{Render}(x_i, \mathbf{f}^{i}(\mathbf{x}_i), \mathbf{d}),
    \label{eq:render}
\end{equation}
where we use an MLP to parameterize the $\mathtt{Render}$ module. The ``interpolated'' surfel feature $\mathbf{f}^{i}(\mathbf{x}_i)$ is given as:
\begin{equation}
     \mathbf{f}^{i}(\mathbf{x}_i) = \frac{\| \mathbf{x}^i - \mathbf{p}^i \|}{r^{i}}\mathtt{F}(\mathbf{f}^{i}, \mathbf{d}, \mathbf{n}^{i}, w^{i}), 
    \label{eq:shading_feats}
\end{equation}
where $\mathtt{F}$ is also parameterized as an  MLP-like network. Note that a recent concurrent work~\cite{zhang2022differentiable} also explores the hybrid rendering process by combining rasterization and volume rendering, but using a different neural representation.




\subsection{Training and Optimization}

All modules in SurfelNeRF are differentiable, so they can be trained end-to-end on large-scale scene datasets with sequential images as input. We utilize L2 pixel-wise rendering loss to train our network supervised by the target ground truth image:
\begin{equation}
	\mathcal{L}_{\mathtt{render}} =  \vert \vert\ C_{{gt}} - C_{{r}} \vert \vert ^2_2,
	\label{eq:loss_color}
\end{equation}
where $C_{{gt}}$ is the ground truth pixel color and $C_{{r}}$ denotes the rendered pixel color output from SurfelNeRF. 
In addition, if we can acquire sensor depth $D^{\mathtt{sen}}_i$ senor for image $I_i$, we further employ an L1 loss to train the depth refinement network:
\begin{equation}
	\mathcal{L}_{d} = \vert \vert\ D^{\mathtt{sen}}_i \odot M_i - D^{i} \odot M_i \vert \vert\ _1 ,
	\label{eq:loss_depth}
\end{equation}
where $D^{i}$ is the output from depth refinement network, and $M_i$ is a binary mask indicating valid sensor depths. This supervision can make the depth refinement network to learn to match the sensor depths.

To sum up, the total loss to train SurfelNeRF end-to-end is given as
\begin{equation}
	\mathcal{L} = \mathcal{L}_{\mathtt{render}} + \lambda \cdot \mathbbm{1}(\mathcal{L}_{d}),
	\label{eq:total_loss}
\end{equation}
where $\mathbbm{1}$ denotes the indicator function whose value is set to 1 if sensor depth is available; $\lambda$ is the balancing weight.

\noindent\textbf{Training Details.}
Following~\cite{zhang2022nerfusion}, we uniformly sample keyframes from each image sequence as input for training SurfelNeRF. To train SurfelNeRF, we sample about 5\% of the full sequence as keyframes, and other frames are used as supervision.

\noindent\textbf{Per-scene Fine-tuning.}
After being trained on large-scale datasets, our SurfelNeRF is capable of constructing neural surfel radiance fields and rendering photorealistic novel views on unseen scenes in a feedforward manner. To further improve the rendering performance on test scenes, one can also perform a \textit{short} period of per-scene optimization via fine-tuning the surfel representations and the rendering module. 
\gao{
To make a fair comparison, we do not use the additional depths in the fine-tuning process and keep the depth refinement module fixed.
}
Thanks to our efficient rendering scheme, SurfelNeRF takes less time when fine-tuning compared with existing methods~\cite{zhang2022nerfusion, xu2022point}.


\section{Experiments}

\subsection{Experimental Settings}

\noindent\textbf{Training Datasets.}
Our training data consists of large-scale indoor scenes from ScanNet~\cite{DBLP:conf/cvpr/DaiCSHFN17} which provides the online sequence input of images with known camera parameters, and the captured incomplete sensor depths. We focus on the online input of large-scale scenes in the main paper.
Following~\cite{zhang2022nerfusion,wang2021ibrnet}, we randomly sample 100 scenes from ScanNet for training.

\noindent\textbf{Implementation Details.}
We train our SurfelNeRF on the large-scale training dataset mentioned above using the Adam~\cite{kingma2014adam} optimizer with an initial learning rate of 0.001 for total 200k iterations on 4 NVIDIA Tesla V100 32GB GPUs. 
For per-scene optimization settings, we fine-tune our method also using the Adam optimizer with an initial learning rate of 0.0002. 
All input images are resized to $640 \times 480$. The $\lambda$ in Eq.~\ref{eq:total_loss} is set to 0.1. The threshold of $\delta_{depth}$ is set to 0.1m in surfel fusion (Sec.~\ref{sec:surfel_fusion}). The maximum number of candidates M in Eq.~\ref{eq:volume_render} for rendering is set to 80. We implement our method using Pytorch and Pytorch3D.


\noindent\textbf{Metrics.}
Following previous works~\cite{zhang2022nerfusion}, we report the PSNR, SSIM~\cite{wang2004image} and $\mathrm{LPIPS}_{vgg}$~\cite{zhang2018unreasonable} to evaluate and compare the results. For fair comparisons with previous methods, we also evaluate our method on the same test set on the ScanNet dataset. The time cost of rendering and per-scene fine-tuning is measured on the same platform using one NVIDIA Tesla V100 32GB GPU.

\subsection{Comparison with state-of-the-arts}

In this section, we evaluate our method with other state-of-the-arts on the large-scale dataset, ScanNet. To achieve fair comparisons, we evaluate our methods strictly following the same training and evaluation scheme as NeRFusion~\cite{zhang2022nerfusion} and NerfingMVS~\cite{wei2021nerfingmvs}. There are 8 selected testing scenes in ScanNet where 1/8 of images are held out for testing. Following~\cite{zhang2022nerfusion}, we report no per-scene optimization results via direct inference network and per-scene optimization results.

\noindent\textbf{Baselines.} In addition to comparing SurfelNeRF with NeRFusion~\cite{zhang2022nerfusion}, we also compare it with other state-of-the-art (SOTA) methods, such as fast per-scene optimization NeRF and neural point cloud representation, generalizable NeRF, \gao{scalable MLP-based representation NeRF, NeRF with additional depth supervision,} and classical image-based rendering methods, including Instant-NGP~\cite{mueller2022instant}, ADOP~\cite{ruckert2022adop}, PointNeRF~\cite{xu2022point}, Mip-NeRF-360~\cite{barron2022mip}, Dense Depth Prior NeRF(DDP-NeRF)~\cite{roessle2022depthpriorsnerf}, and IBRNet~\cite{wang2021ibrnet}. 
\gao{Among these methods, Mip-NeRF-360 is one of the state-of-the-art NeRF methods for large-scale scenes. DDP-NeRF employs an offline NeRF with additional depth supervision.}
PointNeRF achieves state-of-the-art performance thanks to point-based representation. ADOP achieves promising results on offline large-scale rendering. These two methods employ a similar representation to our method; thus we also compare with them. We run generalizable PointNeRF with the same No per-scene optimization and per-scene optimization settings using its open-source code.

The result is shown in Tab.~\ref{tab:com_sota} (a). For no per-scene optimization settings, NeRFusion achieves PSNR of 22.99 thanks to the fusion between input features to global grid representations. Our SurfelNeRF outperforms NeRFusion since SurfelNeRF benefits from compact and precise feature fusion and the geometric guidance of surfels. For the per-scene optimization task, the result is shown in Tab.~\ref{tab:com_sota} (b). PointNeRF gains a large improvement after fine-tuning, which illustrates it achieves sub-optimal performance in no per-scene optimization settings due to its inability to fuse and update point features. 
\gao{Compared with DDP-NeRF, which employs additional depth supervision, our SurfelNeRF still outperforms it by taking advantage of surfel representation and its integration.}
Our method achieves state-of-the-art results again, which shows that our neural surfel representation can provide optimal initialization for fine-tuning.

\begin{figure*}[th]
	\centering
	\begin{minipage}[t]{0.9\linewidth}
		\centering
		\includegraphics[width=1\textwidth]{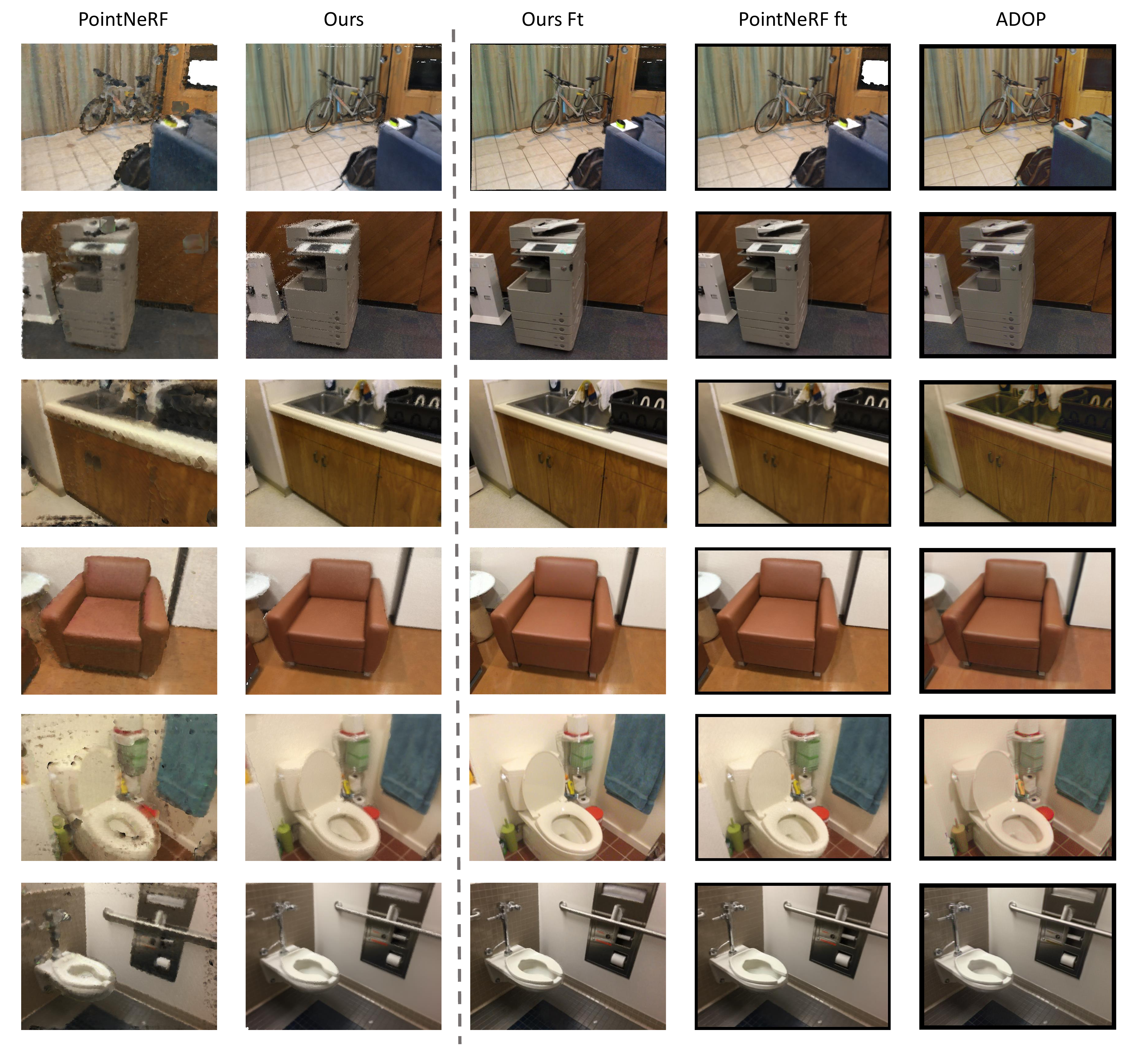}
	\end{minipage}
    \vspace{-5pt}
	\caption{Qualitative comparisons of rendering quality. The 1-2 column are the results of no per-scene optimization. And the left three columns are per-scene optimization results.  It is obvious that with or without per-scene optimization, our method can still achieve the most photorealistic rendering results. Since PointNeRF without a depth refinement network relies on captured depths only, it cannot handle the scenario that depth is incomplete. However, our method can tackle this problem with a lightweight depth refinement network. ADOP can produce geometrically correct results, but its color is less photorealistic.
	}
\label{fig:vis}
\vspace{-8pt}
\end{figure*}

The qualitative comparison is shown in Figure.~\ref{fig:vis}. We compare our method with PointNeRF without per-scene optimization, and compare with per-scene optimization methods of PoingNeRF and ADOP. Since NeRFusion has not release the executable code, we cannot generate qualitative results. As shown in Figure.~\ref{fig:vis}, our results are visually much better than state-of-the-art methods, PointNeRF and ADOP.

\begin{table}[t]
	\center
	\begin{center}
		\center
		\begin{center}
			\setlength{\tabcolsep}{3 pt}
			\begin{tabular}{ccccc}
				\toprule
Methods  & PSNR$\uparrow$ & SSIM$\uparrow$ &LPIPS $\downarrow$ & Time $\downarrow$ \\ \midrule
IBRNet~\cite{wang2021ibrnet}         &21.19  &0.786  &0.358   & - \\ 			   
NeRFusion~\cite{zhang2022nerfusion}           &22.99 &0.838  &0.335 &  $38s^\sharp$ \\ 
PointNeRF~\cite{xu2022point}           & 20.47  &0.642  &0.544    &  30s \\     
SurfelNeRF          &\textbf{23.82}  & \textbf{0.845}  & \textbf{0.327}   & 0.2s \\    \bottomrule
\end{tabular}
\label{tab:no_per_scene}
\subcaption{Results of No per-scene optimization setting on ScanNet. And we report the average time to render an image. $\sharp$ indicates that the time of NeRFusion is reported by authors using a 2080Ti since they have not released executable code before submission.}
\vspace{4 mm}

\begin{tabular}{ccccc}
\toprule
Methods  & PSNR$\uparrow$ & SSIM$\uparrow$ &LPIPS $\downarrow$ & Time $\downarrow$ \\ \midrule
Instant-NGP~\cite{mueller2022instant}           & 23.23 & 0.714  & 0.459 &  0.03  \\ 
ADOP~\cite{ruckert2022adop}           &25.01  &0.807  &0.272   &  1s \\ 
NeRFingMVS~\cite{wei2021nerfingmvs}           &26.37 &0.903  &0.245 &   -  \\
IBRNet~\cite{wang2021ibrnet}          &25.14  &0.871  &0.266   & -  \\
DDP-NeRF~\cite{roessle2022depthpriorsnerf} &26.61  &0.80  &0.26   & - \\
Mip-NeRF-360~\cite{barron2022mip} &27.85  &0.86  &0.25   & - \\
NeRFusion~\cite{zhang2022nerfusion} &26.49  &0.915  &\textbf{0.209}  &  38s \\
PointNeRF~\cite{xu2022point}             &28.99  &0.829  &0.324  & 30s \\ 
SurfelNeRF           &\textbf{29.58}  & \textbf{0.919}  &0.215   & 0.2s \\ 
       \bottomrule
		\end{tabular}
\label{tab:per_scene_opt}
   \subcaption{Per-scene optimization. Time$\downarrow$ indicates the average time to render an image. }
		\end{center}
	\end{center}
	\caption{Quantitative comparisons with SOTAs on the ScanNet dataset with No per-scene and per-scene optimization. }
    \vspace{-12pt}
	\label{tab:com_sota}
\end{table}

\subsection{Ablation Study}

We investigate the effectiveness of designs in our SurfelNeRF by conducting the following ablation studies on ScanNet.

\noindent\textbf{Fusion Schemes.}
Our surfel-based fusion is inspired by classical Surfel-based fusion in conventional 3D reconstruction~\cite{whelan2015elasticfusion}. The classical surfel-based fusion update features via the simple weighted sum of two overlapped surfels. We conduct an experiment that updates surfel features by weighted sum based on surfel weights, which is a similar operation of updating geometry attributes in Eq.~\ref{eq:surfel_update}.
The results are shown in Tab.~\ref{tab:ablation_surfel_fusion}. Since the GRU module can learn to adaptively update features based on the global surfels and new incremental surfels, it can improve performance with negligible computation overhead.

\begin{table}[t]
	\small
	\center
	\begin{center}
		\begin{tabular}{ccccc}
\toprule
Fusion Scheme  & Setting & PSNR$\uparrow$ & SSIM$\uparrow$ &LPIPS $\downarrow$  \\ \midrule
Weighted Sum  & No per-scene       & 23.09 & 0.833  & 0.353   \\ 
GRU           & optimization  &  23.82 & 0.845  & 0.327  \\ 
       \midrule
Weighted Sum  & Per-scene       & 28.54 & 0.884  & 0.293   \\ 
GRU           & optimization  &\textbf{29.58}  & \textbf{0.919}  &\textbf{0.215}  \\ \bottomrule
		\end{tabular}
	\end{center}
	\caption{Ablation studies about fusion schemes in our  SurfelNeRF.}
	\label{tab:ablation_surfel_fusion}
    \vspace{-12pt}
\end{table}

\noindent\textbf{Depth Refinement Module.}
To investigate the effectiveness of our depth refinement, we conduct experiments that reconstruct scenes with incomplete depth captured by sensors. The quantitative results are shown in Tab.~\ref{tab:ablation_depth}, \gao{which show that PSNR performance drops about $4\%$ without the depth refinement. Depth quality affects surfel reconstruction geometry quality, but captured depth is always incomplete (visualized in the first row in Fig.~\ref{fig:vis}). Thus, a depth refinement network is necessarily required to refine the input depth for better surfel geometry quality. 
The experimental result shows that the depth refinement network is effective in alleviating the noisy or incomplete captured or estimated depth.
Moreover, we also conduct an experiment with worse depth quality that resizes input depth resolution to half size. Our performance just sightly drops $0.5\%$ PSNR. These experiments show the effectiveness of our depth refinement module.
}

\begin{table}[t]
	\small
	\center
	\begin{center}
		\begin{tabular}{cccc}
\toprule
 Depth Refinement & PSNR$\uparrow$ & SSIM$\uparrow$ &LPIPS $\downarrow$  \\ \midrule
 w/o      & 19.51 & 0.713  & 0.454   \\ 
 w      & 23.09 & 0.833  & 0.353   \\ 
\bottomrule
		\end{tabular}
	\end{center}
\caption{Ablation studies about the effectiveness of depth refinement in $scene0000\_{01}$ in the ScanNet with the no per-scene optimization setting. 
The depth captured by the sensor in this scene is incomplete. If using captured depths only, some pixels cannot be projected to the 3D space, as the visualization results of PointNeRF shown in the first row in the qualitative results in Fig.~\ref{fig:vis}.
}
\label{tab:ablation_depth}
\end{table}



\noindent\textbf{Rendering Schemes.}
Our proposed renderer renders a ray with a few shading points, which are the intersection points of ray and surfels, unlike volume rendering in NeRF~\cite{mildenhall2020nerf} that requires hundreds of shading points each with one MLP evaluation.
To validate the effectiveness of our differentiable rasterization scheme for rendering, we conduct an experiment that replaced our rendering scheme with NeRFs' volume rendering. The result is shown in Tab.~\ref{tab:ablation_rendering_scheme}.
\gao{
Since online integration of surfels can limit the number of intersecting surfels, our rendering speed is persistently faster than volume rendering with fewer shading points. Specifically, The average and maximal number of rasterized surfels per pixel at novel views on ScanNet are 2.92 and 10, respectively.
}

\begin{table}[t]
	\small
	\center
	\begin{center}
		\begin{tabular}{cccc}
\toprule
Rendering Scheme  & Setting & PSNR$\uparrow$  &Time $\downarrow$  \\ \midrule
Volume Rendering  & No per-scene   & 23.80      &  30s   \\ 
Ours           & optimization  &  23.82  & 0.2s \\ 
       \midrule
Volume Rendering & Per-scene       &  29.45 &   30s   \\ 
Ours          & optimization  &  \textbf{29.58} &    \textbf{0.2s}  \\ \bottomrule
		\end{tabular}
	\end{center}
	\caption{Comparing with two rendering schemes, volume rendering~\cite{mildenhall2020nerf} and our proposed rasterization-guided rendering, in the ScanNet dataset.}
	\label{tab:ablation_rendering_scheme}
 \vspace{-10pt}
\end{table}

\begin{figure}[t]
\centering
\begin{minipage}[t]{0.475\columnwidth}
  \includegraphics[width=\linewidth]{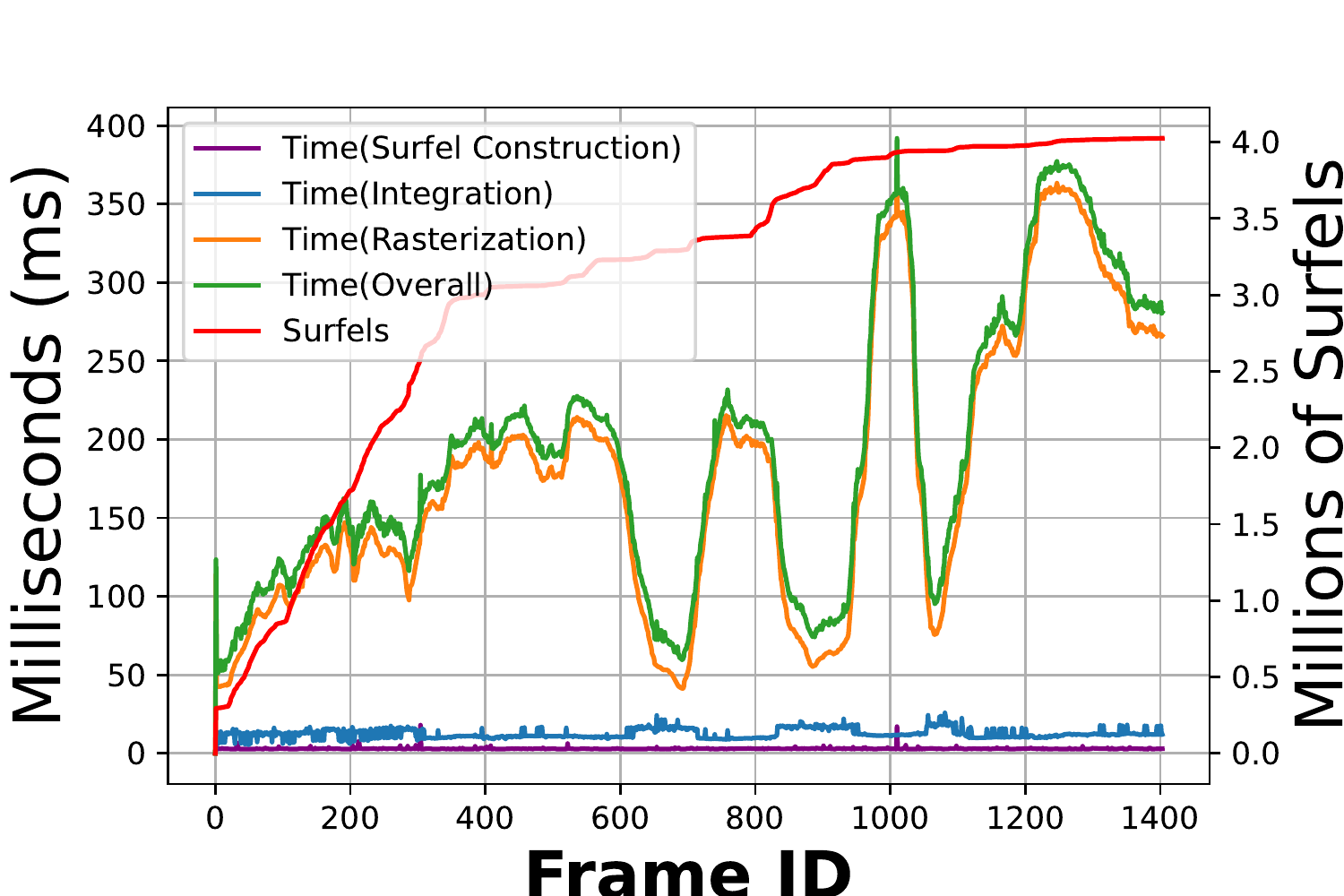}
  \subcaption{GPU processing time.}
  \label{fig:fig_time}
\end{minipage}\hfill 
\begin{minipage}[t]{0.475\columnwidth}
\centering
  \includegraphics[width=\linewidth]{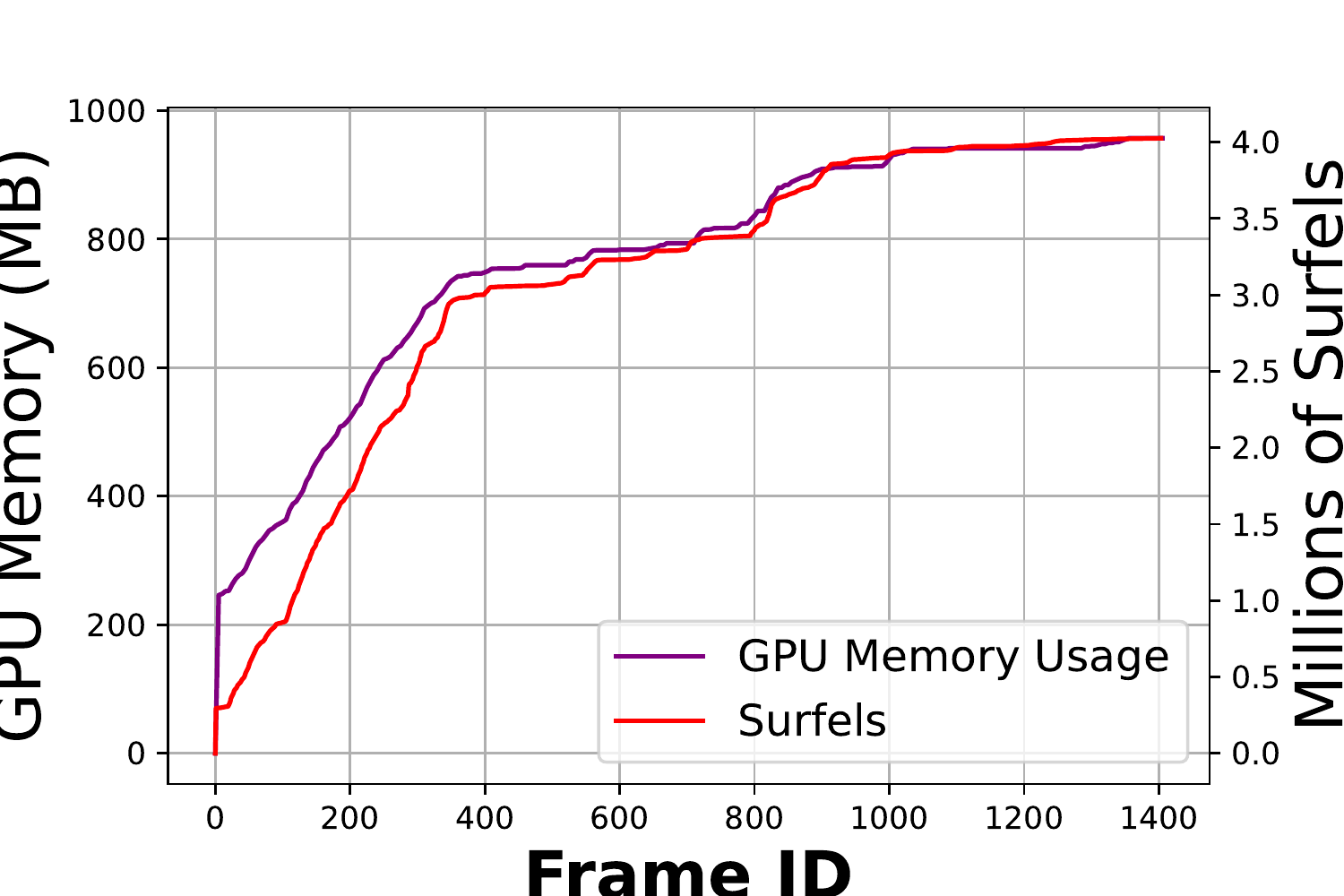}
  \subcaption{GPU memory usage.}
  \label{fig:fig_memory}
\end{minipage}
\caption{The computation cost, memory footprint, and the number of surfels grow according to input frames measured on a three rooms scene.}
\label{fig:fig_computation_and_memory}
\vspace{-10pt}
\end{figure}

\noindent\textbf{Computational Cost and Memory Usage.}
\gao{
To evaluate the computational cost and memory usage on a large-scale scene, we plot components of GPU processing time, surfel
number, and memory usage of our method on a large three rooms scene measured on a V100 GPU implemented by Pytorch3D. During inference, our method renders $640\times 480$ images at 5.2FPS on average. As shown in Fig.~\ref{fig:fig_time} and Fig.~\ref{fig:fig_memory}, the number of surfels has a sub-linear growth thanks to the proposed surfel integration.
The fluctuation of \textit{Time (Rasterization)} is caused by the different numbers of corresponding surfels at novel views.
Although the computational cost of rasterization depends on the number of rasterized surfels, but it is still faster than volume rendering~\cite{mildenhall2020nerf}.
These results show that our surfel representation is compact and efficient. The major computation cost comes from rasterization, and the cost of surfel construction as well as integration could be neglected.
}

\section{Limitations}
\gao{
Our method requires high-quality surfel geometry as neural surfel representation since SurfelNeRF relies on it to conduct surfel integration and rasterization-guided render. To obtain better geometry quality, we may require more accurate depth quality to construct surfel geometry, which necessitates extra depth supervision or the MVS depth estimator. Since captured or estimated depth quality is not exact, there may be geometry discontinuities or missing parts in the surfel representation. Consequently, there may be some missing parts or incorrect geometry in the rendered output (examples of missing parts are shown in the first row in Fig.~\ref{fig:vis} or video). As discussed in ablation studies of depth refinement above, inaccurate and noisy estimated depth as well as depth discontinuity could affect the construction of neural surfel representation. To alleviate this issue, we propose a depth refinement module (Tab.~\ref{tab:ablation_depth}) to refine depth information.
%
}

\section{Conclusion}
In this paper, we propose SurfelNeRF for online photorealistic reconstruction and rendering of large-scale indoor scenes. We employ neural surfels to store a neural radiance field, adopt an efficient neural radiance field fusion method to progressively build the scene representation, and propose a fast and differentiable rasterization-based rendering algorithm to tackle the challenging task. Experiments on the ScanNet dataset demonstrate that our method achieving favorably better
results than the state-of-the-art methods.



{\small
\bibliographystyle{ieee_fullname}
\bibliography{egbib}

\begin{thebibliography}{10}\itemsep=-1pt

\bibitem{npbg}
Kara-Ali Aliev, Artem Sevastopolsky, Maria Kolos, Dmitry Ulyanov, and Victor
  Lempitsky.
\newblock Neural point-based graphics.
\newblock 2020.

\bibitem{barron2021mip}
Jonathan~T Barron, Ben Mildenhall, Matthew Tancik, Peter Hedman, Ricardo
  Martin-Brualla, and Pratul~P Srinivasan.
\newblock Mip-nerf: A multiscale representation for anti-aliasing neural
  radiance fields.
\newblock In {\em Proceedings of the IEEE/CVF International Conference on
  Computer Vision}, pages 5855--5864, 2021.

\bibitem{barron2022mip}
Jonathan~T Barron, Ben Mildenhall, Dor Verbin, Pratul~P Srinivasan, and Peter
  Hedman.
\newblock Mip-nerf 360: Unbounded anti-aliased neural radiance fields.
\newblock In {\em Proceedings of the IEEE/CVF Conference on Computer Vision and
  Pattern Recognition}, pages 5470--5479, 2022.

\bibitem{bozic2021transformerfusion}
Aljaz Bozic, Pablo Palafox, Justus Thies, Angela Dai, and Matthias Nie{\ss}ner.
\newblock Transformerfusion: Monocular rgb scene reconstruction using
  transformers.
\newblock {\em Advances in Neural Information Processing Systems},
  34:1403--1414, 2021.

\bibitem{cao2022fwd}
Ang Cao, Chris Rockwell, and Justin Johnson.
\newblock Fwd: Real-time novel view synthesis with forward warping and depth.
\newblock In {\em Proceedings of the IEEE/CVF Conference on Computer Vision and
  Pattern Recognition}, pages 15713--15724, 2022.

\bibitem{Cao_TOG_2018}
Yan-Pei Cao, Leif Kobbelt, and Shi-Min Hu.
\newblock Real-time high-accuracy three-dimensional reconstruction with
  consumer {RGB-D} cameras.
\newblock {\em ACM Transactions on Graphics}, 37(5):171:1--171:16, 2018.

\bibitem{chen2022tensorf}
Anpei Chen, Zexiang Xu, Andreas Geiger, Jingyi Yu, and Hao Su.
\newblock Tensorf: Tensorial radiance fields.
\newblock {\em arXiv preprint arXiv:2203.09517}, 2022.

\bibitem{chen2021mvsnerf}
Anpei Chen, Zexiang Xu, Fuqiang Zhao, Xiaoshuai Zhang, Fanbo Xiang, Jingyi Yu,
  and Hao Su.
\newblock Mvsnerf: Fast generalizable radiance field reconstruction from
  multi-view stereo.
\newblock In {\em Proceedings of the IEEE/CVF International Conference on
  Computer Vision}, pages 14124--14133, 2021.

\bibitem{chen2022mobilenerf}
Zhiqin Chen, Thomas Funkhouser, Peter Hedman, and Andrea Tagliasacchi.
\newblock Mobilenerf: Exploiting the polygon rasterization pipeline for
  efficient neural field rendering on mobile architectures.
\newblock {\em arXiv preprint arXiv:2208.00277}, 2022.

\bibitem{clark2022volumetric}
Ronald Clark.
\newblock Volumetric bundle adjustment for online photorealistic scene capture.
\newblock In {\em Proceedings of the IEEE/CVF Conference on Computer Vision and
  Pattern Recognition}, pages 6124--6132, 2022.

\bibitem{DBLP:conf/cvpr/DaiCSHFN17}
Angela Dai, Angel~X. Chang, Manolis Savva, Maciej Halber, Thomas~A. Funkhouser,
  and Matthias Nie{\ss}ner.
\newblock Scannet: Richly-annotated 3d reconstructions of indoor scenes.
\newblock In {\em {IEEE} {CVPR}}, pages 2432--2443, 2017.

\bibitem{BundleFusion}
Angela Dai, Matthias Nie{\ss}ner, Michael Zollh{\"{o}}fer, Shahram Izadi, and
  Christian Theobalt.
\newblock Bundlefusion: Real-time globally consistent 3d reconstruction using
  on-the-fly surface reintegration.
\newblock {\em ACM Transactions on Graphics}, 36(3):24:1--24:18, 2017.

\bibitem{hedman2021baking}
Peter Hedman, Pratul~P Srinivasan, Ben Mildenhall, Jonathan~T Barron, and Paul
  Debevec.
\newblock Baking neural radiance fields for real-time view synthesis.
\newblock In {\em Proceedings of the IEEE/CVF International Conference on
  Computer Vision}, pages 5875--5884, 2021.

\bibitem{ReluField_sigg_22}
Animesh Karnewar, Tobias Ritschel, Oliver Wang, and Niloy Mitra.
\newblock Relu fields: The little non-linearity that could.
\newblock In {\em ACM SIGGRAPH 2022 Conference Proceedings}, SIGGRAPH '22, New
  York, NY, USA, 2022. Association for Computing Machinery.

\bibitem{keller2013real}
Maik Keller, Damien Lefloch, Martin Lambers, Shahram Izadi, Tim Weyrich, and
  Andreas Kolb.
\newblock Real-time 3d reconstruction in dynamic scenes using point-based
  fusion.
\newblock In {\em 2013 International Conference on 3D Vision-3DV 2013}, pages
  1--8. IEEE, 2013.

\bibitem{10.1145/3503926}
Jungeon Kim, Hyomin Kim, Hyeonseo Nam, Jaesik Park, and Seungyong Lee.
\newblock Textureme: High-quality textured scene reconstruction in real time.
\newblock {\em ACM Trans. Graph.}, 41(3), mar 2022.

\bibitem{kingma2014adam}
Diederik~P Kingma and Jimmy Ba.
\newblock Adam: A method for stochastic optimization.
\newblock {\em arXiv preprint arXiv:1412.6980}, 2014.

\bibitem{li2021neulf}
Zhong Li, Liangchen Song, Celong Liu, Junsong Yuan, and Yi Xu.
\newblock Neulf: Efficient novel view synthesis with neural 4d light field.
\newblock {\em arXiv preprint arXiv:2105.07112}, 2021.

\bibitem{liu2022devrf}
Jia-Wei Liu, Yan-Pei Cao, Weijia Mao, Wenqiao Zhang, David~Junhao Zhang, Jussi
  Keppo, Ying Shan, Xiaohu Qie, and Mike~Zheng Shou.
\newblock Devrf: Fast deformable voxel radiance fields for dynamic scenes.
\newblock {\em arXiv preprint arXiv:2205.15723}, 2022.

\bibitem{liu2020neural}
Lingjie Liu, Jiatao Gu, Kyaw~Zaw Lin, Tat-Seng Chua, and Christian Theobalt.
\newblock Neural sparse voxel fields.
\newblock {\em NeurIPS}, 2020.

\bibitem{mihajlovic2021deepsurfels}
Marko Mihajlovic, Silvan Weder, Marc Pollefeys, and Martin~R Oswald.
\newblock Deepsurfels: Learning online appearance fusion.
\newblock In {\em Proceedings of the IEEE/CVF Conference on Computer Vision and
  Pattern Recognition}, pages 14524--14535, 2021.

\bibitem{mildenhall2020nerf}
Ben Mildenhall, Pratul~P Srinivasan, Matthew Tancik, Jonathan~T Barron, Ravi
  Ramamoorthi, and Ren Ng.
\newblock Nerf: Representing scenes as neural radiance fields for view
  synthesis.
\newblock In {\em European conference on computer vision}, pages 405--421.
  Springer, 2020.

\bibitem{mueller2022instant}
Thomas M\"uller, Alex Evans, Christoph Schied, and Alexander Keller.
\newblock Instant neural graphics primitives with a multiresolution hash
  encoding.
\newblock {\em ACM Trans. Graph.}, 41(4):102:1--102:15, July 2022.

\bibitem{KinectFusion}
Richard~A. Newcombe, Shahram Izadi, Otmar Hilliges, David Molyneaux, David Kim,
  Andrew~J. Davison, Pushmeet Kohli, Jamie Shotton, Steve Hodges, and Andrew~W.
  Fitzgibbon.
\newblock Kinectfusion: Real-time dense surface mapping and tracking.
\newblock In {\em IEEE ISMAR}, pages 127--136, 2011.

\bibitem{pfister2000surfels}
Hanspeter Pfister, Matthias Zwicker, Jeroen Van~Baar, and Markus Gross.
\newblock Surfels: Surface elements as rendering primitives.
\newblock In {\em Proceedings of the 27th annual conference on Computer
  graphics and interactive techniques}, pages 335--342, 2000.

\bibitem{Rakhimov_2022_CVPR}
Ruslan Rakhimov, Andrei-Timotei Ardelean, Victor Lempitsky, and Evgeny Burnaev.
\newblock Npbg++: Accelerating neural point-based graphics.
\newblock In {\em Proceedings of the IEEE/CVF Conference on Computer Vision and
  Pattern Recognition (CVPR)}, pages 15969--15979, June 2022.

\bibitem{rematas2021urban}
Konstantinos Rematas, Andrew Liu, Pratul~P Srinivasan, Jonathan~T Barron,
  Andrea Tagliasacchi, Thomas Funkhouser, and Vittorio Ferrari.
\newblock Urban radiance fields.
\newblock {\em arXiv preprint arXiv:2111.14643}, 2021.

\bibitem{roessle2022depthpriorsnerf}
Barbara Roessle, Jonathan~T. Barron, Ben Mildenhall, Pratul~P. Srinivasan, and
  Matthias Nie{\ss}ner.
\newblock Dense depth priors for neural radiance fields from sparse input
  views.
\newblock In {\em Proceedings of the IEEE/CVF Conference on Computer Vision and
  Pattern Recognition (CVPR)}, June 2022.

\bibitem{ronneberger2015u}
Olaf Ronneberger, Philipp Fischer, and Thomas Brox.
\newblock U-net: Convolutional networks for biomedical image segmentation.
\newblock In {\em International Conference on Medical image computing and
  computer-assisted intervention}, pages 234--241. Springer, 2015.

\bibitem{ruckert2021adop}
Darius R{\"u}ckert, Linus Franke, and Marc Stamminger.
\newblock Adop: Approximate differentiable one-pixel point rendering.
\newblock {\em arXiv preprint arXiv:2110.06635}, 2021.

\bibitem{ruckert2022adop}
Darius R{\"u}ckert, Linus Franke, and Marc Stamminger.
\newblock Adop: Approximate differentiable one-pixel point rendering.
\newblock {\em ACM Transactions on Graphics (TOG)}, 41(4):1--14, 2022.

\bibitem{sajjadi2022scene}
Mehdi~SM Sajjadi, Henning Meyer, Etienne Pot, Urs Bergmann, Klaus Greff, Noha
  Radwan, Suhani Vora, Mario Lu{\v{c}}i{\'c}, Daniel Duckworth, Alexey
  Dosovitskiy, et~al.
\newblock Scene representation transformer: Geometry-free novel view synthesis
  through set-latent scene representations.
\newblock In {\em Proceedings of the IEEE/CVF Conference on Computer Vision and
  Pattern Recognition}, pages 6229--6238, 2022.

\bibitem{sayed2022simplerecon}
Mohamed Sayed, John Gibson, Jamie Watson, Victor Prisacariu, Michael Firman,
  and Cl{\'e}ment Godard.
\newblock Simplerecon: 3d reconstruction without 3d convolutions.
\newblock In {\em Proceedings of the European Conference on Computer Vision
  (ECCV)}, 2022.

\bibitem{sun2021direct}
Cheng Sun, Min Sun, and Hwann-Tzong Chen.
\newblock Direct voxel grid optimization: Super-fast convergence for radiance
  fields reconstruction.
\newblock {\em arXiv preprint arXiv:2111.11215}, 2021.

\bibitem{NeuralRecon}
Jiaming Sun, Yiming Xie, Linghao Chen, Xiaowei Zhou, and Hujun Bao.
\newblock Neuralrecon: Real-time coherent 3d reconstruction from monocular
  video.
\newblock In {\em {IEEE} {CVPR}}, pages 15598--15607, 2021.

\bibitem{tan2019mnasnet}
Mingxing Tan, Bo Chen, Ruoming Pang, Vijay Vasudevan, Mark Sandler, Andrew
  Howard, and Quoc~V Le.
\newblock Mnasnet: Platform-aware neural architecture search for mobile.
\newblock In {\em Proceedings of the IEEE/CVF Conference on Computer Vision and
  Pattern Recognition}, pages 2820--2828, 2019.

\bibitem{tancik2022block}
Matthew Tancik, Vincent Casser, Xinchen Yan, Sabeek Pradhan, Ben Mildenhall,
  Pratul~P Srinivasan, Jonathan~T Barron, and Henrik Kretzschmar.
\newblock Block-nerf: Scalable large scene neural view synthesis.
\newblock {\em arXiv preprint arXiv:2202.05263}, 2022.

\bibitem{thies2019deferred}
Justus Thies, Michael Zollh{\"o}fer, and Matthias Nie{\ss}ner.
\newblock Deferred neural rendering: Image synthesis using neural textures.
\newblock {\em ACM Transactions on Graphics (TOG)}, 38(4):1--12, 2019.

\bibitem{wang2021ibrnet}
Qianqian Wang, Zhicheng Wang, Kyle Genova, Pratul~P Srinivasan, Howard Zhou,
  Jonathan~T Barron, Ricardo Martin-Brualla, Noah Snavely, and Thomas
  Funkhouser.
\newblock Ibrnet: Learning multi-view image-based rendering.
\newblock In {\em Proceedings of the IEEE/CVF Conference on Computer Vision and
  Pattern Recognition}, pages 4690--4699, 2021.

\bibitem{wang2004image}
Zhou Wang, Alan~C Bovik, Hamid~R Sheikh, and Eero~P Simoncelli.
\newblock Image quality assessment: from error visibility to structural
  similarity.
\newblock {\em IEEE transactions on image processing}, 13(4):600--612, 2004.

\bibitem{weder2020routedfusion}
Silvan Weder, Johannes Schonberger, Marc Pollefeys, and Martin~R Oswald.
\newblock Routedfusion: Learning real-time depth map fusion.
\newblock In {\em Proceedings of the IEEE/CVF Conference on Computer Vision and
  Pattern Recognition}, pages 4887--4897, 2020.

\bibitem{wei2021nerfingmvs}
Yi Wei, Shaohui Liu, Yongming Rao, Wang Zhao, Jiwen Lu, and Jie Zhou.
\newblock Nerfingmvs: Guided optimization of neural radiance fields for indoor
  multi-view stereo.
\newblock In {\em Proceedings of the IEEE/CVF International Conference on
  Computer Vision}, pages 5610--5619, 2021.

\bibitem{whelan2015elasticfusion}
Thomas Whelan, Stefan Leutenegger, Renato Salas-Moreno, Ben Glocker, and Andrew
  Davison.
\newblock Elasticfusion: Dense slam without a pose graph.
\newblock Robotics: Science and Systems, 2015.

\bibitem{xiangli2021citynerf}
Yuanbo Xiangli, Linning Xu, Xingang Pan, Nanxuan Zhao, Anyi Rao, Christian
  Theobalt, Bo Dai, and Dahua Lin.
\newblock Citynerf: Building nerf at city scale.
\newblock {\em arXiv preprint arXiv:2112.05504}, 2021.

\bibitem{xu2022point}
Qiangeng Xu, Zexiang Xu, Julien Philip, Sai Bi, Zhixin Shu, Kalyan Sunkavalli,
  and Ulrich Neumann.
\newblock Point-nerf: Point-based neural radiance fields.
\newblock In {\em Proceedings of the IEEE/CVF Conference on Computer Vision and
  Pattern Recognition}, pages 5438--5448, 2022.

\bibitem{yao2018mvsnet}
Yao Yao, Zixin Luo, Shiwei Li, Tian Fang, and Long Quan.
\newblock Mvsnet: Depth inference for unstructured multi-view stereo.
\newblock {\em European Conference on Computer Vision (ECCV)}, 2018.

\bibitem{yu2021plenoctrees}
Alex Yu, Ruilong Li, Matthew Tancik, Hao Li, Ren Ng, and Angjoo Kanazawa.
\newblock Plenoctrees for real-time rendering of neural radiance fields.
\newblock In {\em Proceedings of the IEEE/CVF International Conference on
  Computer Vision}, pages 5752--5761, 2021.

\bibitem{yu2021pixelnerf}
Alex Yu, Vickie Ye, Matthew Tancik, and Angjoo Kanazawa.
\newblock pixelnerf: Neural radiance fields from one or few images.
\newblock In {\em Proceedings of the IEEE/CVF Conference on Computer Vision and
  Pattern Recognition}, pages 4578--4587, 2021.

\bibitem{zhang2022digging}
Jian Zhang, Jinchi Huang, Bowen Cai, Huan Fu, Mingming Gong, Chaohui Wang,
  Jiaming Wang, Hongchen Luo, Rongfei Jia, Binqiang Zhao, et~al.
\newblock Digging into radiance grid for real-time view synthesis with detail
  preservation.
\newblock In {\em European Conference on Computer Vision}, pages 724--740.
  Springer, 2022.

\bibitem{zhang2022differentiable}
Qiang Zhang, Seung-Hwan Baek, Szymon Rusinkiewicz, and Felix Heide.
\newblock Differentiable point-based radiance fields for efficient view
  synthesis.
\newblock {\em arXiv preprint arXiv:2205.14330}, 2022.

\bibitem{zhang2018unreasonable}
Richard Zhang, Phillip Isola, Alexei~A Efros, Eli Shechtman, and Oliver Wang.
\newblock The unreasonable effectiveness of deep features as a perceptual
  metric.
\newblock In {\em Proceedings of the IEEE conference on computer vision and
  pattern recognition}, pages 586--595, 2018.

\bibitem{zhang2022nerfusion}
Xiaoshuai Zhang, Sai Bi, Kalyan Sunkavalli, Hao Su, and Zexiang Xu.
\newblock Nerfusion: Fusing radiance fields for large-scale scene
  reconstruction.
\newblock In {\em Proceedings of the IEEE/CVF Conference on Computer Vision and
  Pattern Recognition}, pages 5449--5458, 2022.

\bibitem{zhou2018stereo}
Tinghui Zhou, Richard Tucker, John Flynn, Graham Fyffe, and Noah Snavely.
\newblock Stereo magnification: Learning view synthesis using multiplane
  images.
\newblock {\em arXiv preprint arXiv:1805.09817}, 2018.

\bibitem{Zhu2022CVPR}
Zihan Zhu, Songyou Peng, Viktor Larsson, Weiwei Xu, Hujun Bao, Zhaopeng Cui,
  Martin~R. Oswald, and Marc Pollefeys.
\newblock Nice-slam: Neural implicit scalable encoding for slam.
\newblock In {\em IEEE CVPR}, pages 12786--12796, 2022.

\bibitem{zou2022mononeuralfusion}
Zi-Xin Zou, Shi-Sheng Huang, Yan-Pei Cao, Tai-Jiang Mu, Ying Shan, and Hongbo
  Fu.
\newblock Mononeuralfusion: Online monocular neural 3d reconstruction with
  geometric priors.
\newblock {\em arXiv preprint arXiv:2209.15153}, 2022.

\end{thebibliography}
}

\clearpage
\newpage

\renewcommand\thesection{\Alph{section}}
\setcounter{section}{0}

\setcounter{figure}{0} \renewcommand{\thefigure}{A.\arabic{figure}}
\setcounter{table}{0} \renewcommand{\thetable}{A.\arabic{table}}

%

\section{Implementation Details}
\label{sec:details}

\subsection{Network Details}

\noindent\textbf{Image feature extractor.}
We follow~\cite{zhang2022nerfusion} to use the same modified MnasNet~\cite{tan2019mnasnet} pretrained from ImageNet as a 2D CNN to extract surfel features from images. For each surfel,  we extract multi-scale image features from corresponding pixels. The channel number of extracted image features is 83. We then project extracted image features to surfel features with channel number of 32 by an MLP.

\noindent\textbf{GRU fusion network.} We employ a one layer GRU network to fuse the surfel features. Given features $\mathbf{f}_t^{\mathtt{merge}}$  of input surfels and features $\mathbf{f}_{t-1}^{\mathtt{corrs}}$ of corresponding global surfels, the process of updating global surfel features with GRU can be given as:
\begin{equation}
	\mathbf{f}_{t}^{\mathtt{corrs}} = \mathtt{GRU}(\mathbf{f}_t^{\mathtt{merge}}, \mathbf{f}_{t-1}^{\mathtt{corrs}}),
	\label{eq:gru}
\end{equation}
where the detail is expresses by
\begin{equation}
	\begin{split}
	\mathbf{z}_t &= M_z([\mathbf{f}_t^{\mathtt{merge}}, \mathbf{f}_{t-1}^{\mathtt{corrs}}]), \\
	\mathbf{r}_t &= M_r([\mathbf{f}_t^{\mathtt{merge}}, \mathbf{f}_{t-1}^{\mathtt{corrs}}]),  \\
	\tilde{\mathbf{f}}_{t}^{\mathtt{corrs}}  &= M_t([\mathbf{r}_t*\mathbf{f}_{t-1}^{\mathtt{corrs}},\mathbf{f}_t^{\mathtt{merge}}]), \\
	\mathbf{f}_{t}^{\mathtt{corrs}} &= (1-\mathbf{z}_t)*\mathbf{f}_{t-1}^{\mathtt{corrs}} + \mathbf{z}_t*\tilde{\mathbf{f}}_{t}^{\mathtt{corrs}},
	\end{split}
\end{equation}
where $M_z$, $M_r$ and $M_t$ both have one MLP layer followed by a sigmoid, sidmoid and tanh activation function, respectively. $[\cdot, \cdot]$ denotes the operation of concatenate.

\noindent\textbf{Rendering module.} We employ an MLP-like rendering module, $\mathtt{Render}(x_i, \mathbf{f}^{i}(\mathbf{x}_i), \mathbf{d})$, to predict volume density $\sigma_i$ and radiance $c_i$ at each shading point $x_i$ with giving ``interpolated'' surfel features $\mathbf{f}^{i}(\mathbf{x}_i)$ and its view direction $\mathbf{d}$. Specifically, the details can be given as
\begin{equation}
	\begin{split}
		\sigma_i &= F_{\sigma}([\mathbf{f}^{i}(\mathbf{x}_i), \gamma(\mathbf{x}_i)]), \\
		c_i &= \mathrm{Sigmoid}(F_{r}([\mathbf{f}^{i}(\mathbf{x}_i), \gamma(\mathbf{d})])), \\
	\end{split}
\end{equation}
where $F_{\sigma}(\cdot)$ is a one layer MLP network with the ReLU activation function. $F_{r}(\cdot)$ is an  MLP network with four layers and ReLU activation function, where the channel number of all hidden layers is 256. $\gamma(\cdot)$ denotes the positional embedding with maximum frequency of 5. $\mathbf{x}_i$ is the position of shading point. $[\cdot, \cdot]$ is the concatenation operation.
The ``interpolated'' surfel features $\mathbf{f}^{i}(\mathbf{x}_i)$ are obtained based on the intersection of ray and surfels as 
\begin{equation}
	\mathbf{f}^{i}(\mathbf{x}_i) = \frac{r^i - \| \mathbf{x}^i - \mathbf{p}^i \|}{r^{i}}\mathtt{F}(\mathbf{f}^{i}, \mathbf{d}, \mathbf{n}^{i}, w^{i}), 
	\label{eq:shading_feats}
\end{equation}
where $\mathbf{p}^i$, $\mathbf{f}^{i}$, $\mathbf{n}^{i}$, $r^{i}$ and $w^{i}$ indicate the position, features, normal, radius and weight of surfels $s^i$ respectively. The function $\mathtt{F}$ is a MLP-like network, which is given as 
\begin{equation}
	\begin{split}
	\mathtt{F}(\mathbf{f}^{i}, \mathbf{d}, \mathbf{n}^{i}, w^{i}) &= \\ 
	F_{f}([\mathbf{f}^{i}, \gamma(\mathbf{d})&,\gamma(w^{i}), \gamma(\mathbf{n}^{i}),  \gamma(\mathbf{d}-\mathbf{n})]),
	\end{split}
	\label{eq:shading_feats1}
\end{equation}
where $F_f$ is a two layer MLP network with ReLU activation function and the channel number of hidden layers is 256. 

Overall, the function $\mathtt{F}$ takes features of surfels and corresponding geometry attributes of surfels and rays as input, and outputs view-dependent surfel features. 
The view-dependent surfel features are then weighted based on the radius and the distance between intersections and centers of surfels.
The far the intersections are, the less they contribute to the interpolated features.

\section{Additional Results}

\begin{table}[t]
	\center
	\begin{center}
		\setlength{\tabcolsep}{3 pt}
		\center
\begin{tabular}{ccccc}
	\toprule
	Methods  & PSNR$\uparrow$ & SSIM$\uparrow$ &LPIPS $\downarrow$ & Time $\downarrow$ \\ \midrule
	Instant-NGP~\cite{mueller2022instant}           & 23.23 & 0.714  & 0.459 &  0.03s  \\ 
	ADOP~\cite{ruckert2022adop}           &25.01  &0.807  &0.272   &  1s \\ 
	NeRFingMVS~\cite{wei2021nerfingmvs}           &26.37 &0.903  &0.245 &   -  \\
	IBRNet~\cite{wang2021ibrnet}          &25.14  &0.871  &0.266   & -  \\
	NeRFusion~\cite{zhang2022nerfusion}           &26.49  &0.915  &\textbf{0.209}  &  38s \\
	PointNeRF~\cite{xu2022point}           &28.99  &0.829  &0.324  & 30s \\ \midrule
	SurfelNeRF           &29.58  & 0.919  &0.215   & 0.2s \\  
	SurfelNeRF (MVS)          &\textbf{29.74}  & \textbf{0.920}  &0.211   & 0.2s \\ 
	\bottomrule
\end{tabular}
\label{tab:per_scene_opt}
\end{center}
\caption{Quantitative comparisons with SOTAs on the ScanNet dataset with per-scene optimization. SurfelNeRF (MVS) indicates taking RGB input with estimated depth via the MVS depth estimator.
Time$\downarrow$ indicate average time to render an image. }
\label{tab:com_sota}
\end{table}	

\begin{table}[t]
	\center
	\begin{center}
		\setlength{\tabcolsep}{3 pt}
		\center
\begin{tabular}{cccc}
	\toprule
Methods  & PSNR$\uparrow$ & SSIM$\uparrow$ &LPIPS $\downarrow$ \\ \midrule
IBRNet~\cite{wang2021ibrnet}         &21.19  &0.786  &0.358    \\ 			   
NeRFusion~\cite{zhang2022nerfusion}           &22.99 &0.838  &0.335  \\ 
PointNeRF~\cite{xu2022point}           & 20.47  &0.642  &0.544    \\     \midrule
SurfelNeRF          &23.82  & 0.845  & 0.327    \\
SurfelNeRF (MVS)          &\textbf{24.29}  & \textbf{0.871}  & \textbf{0.324}   \\
\bottomrule
\end{tabular}
\label{tab:per_scene_opt}
\end{center}
\caption{Quantitative comparisons with SOTAs on the ScanNet dataset with no per-scene optimization. SurfelNeRF (MVS) indicates taking RGB input with estimated depth via the MVS depth estimator.%
}
\label{tab:com_sota_pretrained}
\end{table}

In this section, we report and analyse quantitative and qualitative results of additional ablation studies, and provide additional qualitative comparison with recent SOTA methods.

\subsection{Depth from MVS}
\label{sec:more_ablation}

We conduct an additional experiment that takes the RGB input from sensors only and employ a off-the-shelf depth estimator~\cite{sayed2022simplerecon} to obtain estimated MVS depth maps. 
We show the quantitative results of this setting with direct network inference and per-scene optimization in Table.~\ref{tab:com_sota_pretrained} and Table.~\ref{tab:com_sota}, respectively.
Our method with estimated depth, called SurfelNeRF(MVS), achieve comparable performance with using sensor depth and a depth refinement network. The slight improvement comes from the scenes where depth captured from the sensor appears heavily incomplete and noisy, where the off-the-shelf depth estimator~\cite{sayed2022simplerecon} can produce better depth via multi-view stereo than raw sensor measurements.
To investigate the influence of depth quality, we conduct an additional ablation study about depth refinement network in the next section.

\subsection{Additional Ablation Studies}
\label{sec:vis_ablation}

\begin{figure*}
	\begin{minipage}[t]{1\linewidth}
		\centering
		\includegraphics[width=1.0\textwidth]{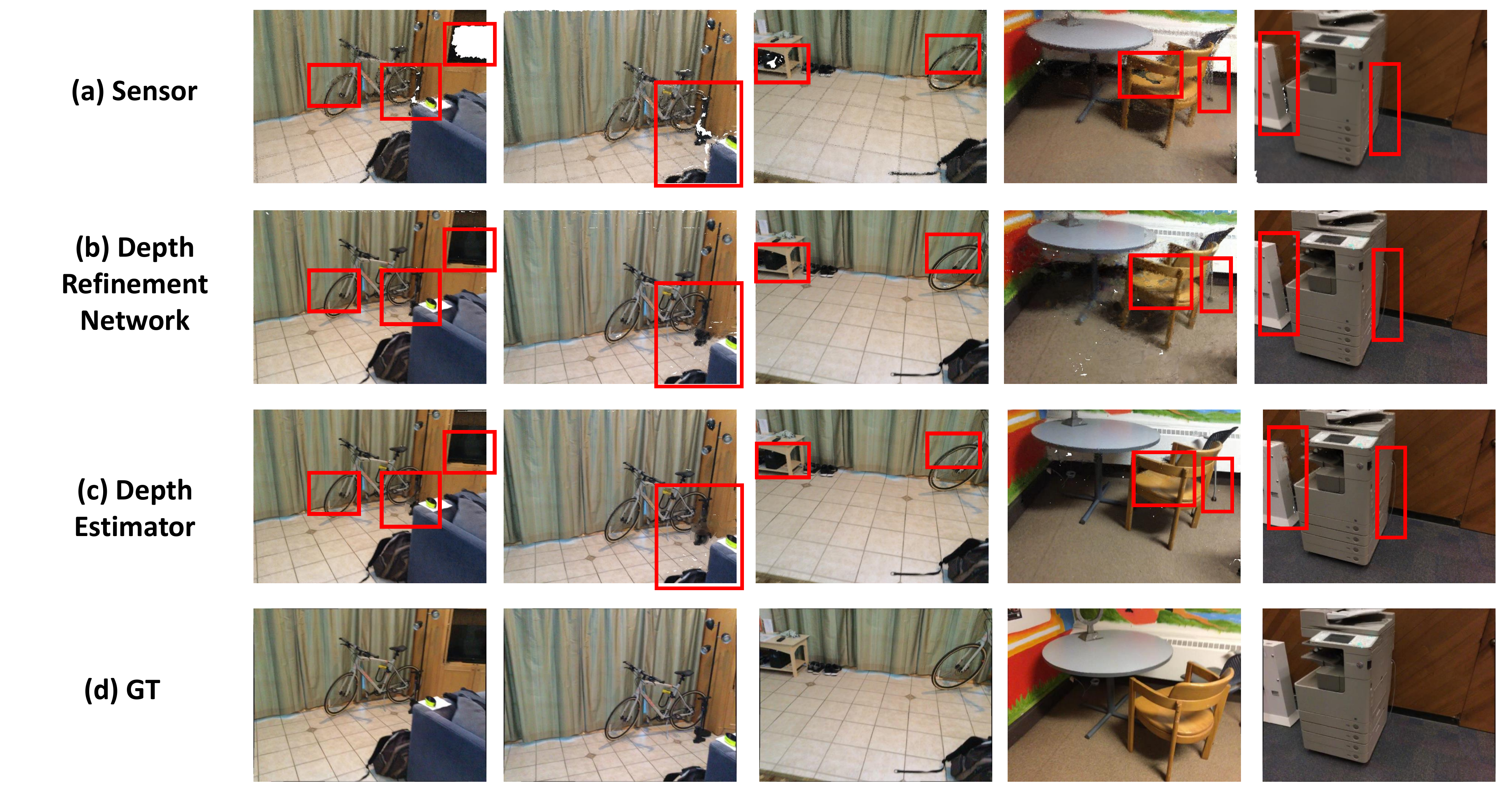}
	\end{minipage}
	\caption{Comparison of different types of input depth on the per-scene optimization setting. Per-sceen optimizaiton would not change the surfel position and number, so it can obtain the same conclusion when evaluating on the no per-scene optimization setting. The highlight areas are indicated by red rectangles. It is obvious that depths captured from sensor may be incomplete and noisy, which affects the surfel reconstruction resulting in sub-optimal rendering results.}
	\label{fig:depth}
\end{figure*}

\noindent\textbf{Depth refinement network.}
We show the visualization results of heavily incomplete scenes with different input depth, including depth captured from sensors, refined from the depth refinement network, and estimated by the off-the-shelf depth estimator, as shown in Figure.~\ref{fig:depth}.
For the first column that a novel view from $scene0000\_01$ in ScanNet, the sensor cannot capture the high-quality depth around the thin bicycle wheels and the far away television. With the help of RGB input or multi-view stereo techniques, the depth refinement network and depth estimator can fill the depth, which reconstruct surfels better and providing better rendering results. Comparing with the depth refinement network, the off-the-shelf depth estimator produce higher quality of estimated depth since it spends extra time to consider the prior of multi-view stereo. Comparing the results with different depth quality, this results shows that the higher quality of depth the better photo-realistic rendering results since depth quality decides the quality of reconstruction surfels.

\begin{figure*}
	\begin{minipage}[t]{1\linewidth}
		\centering
		\includegraphics[width=1.0\textwidth]{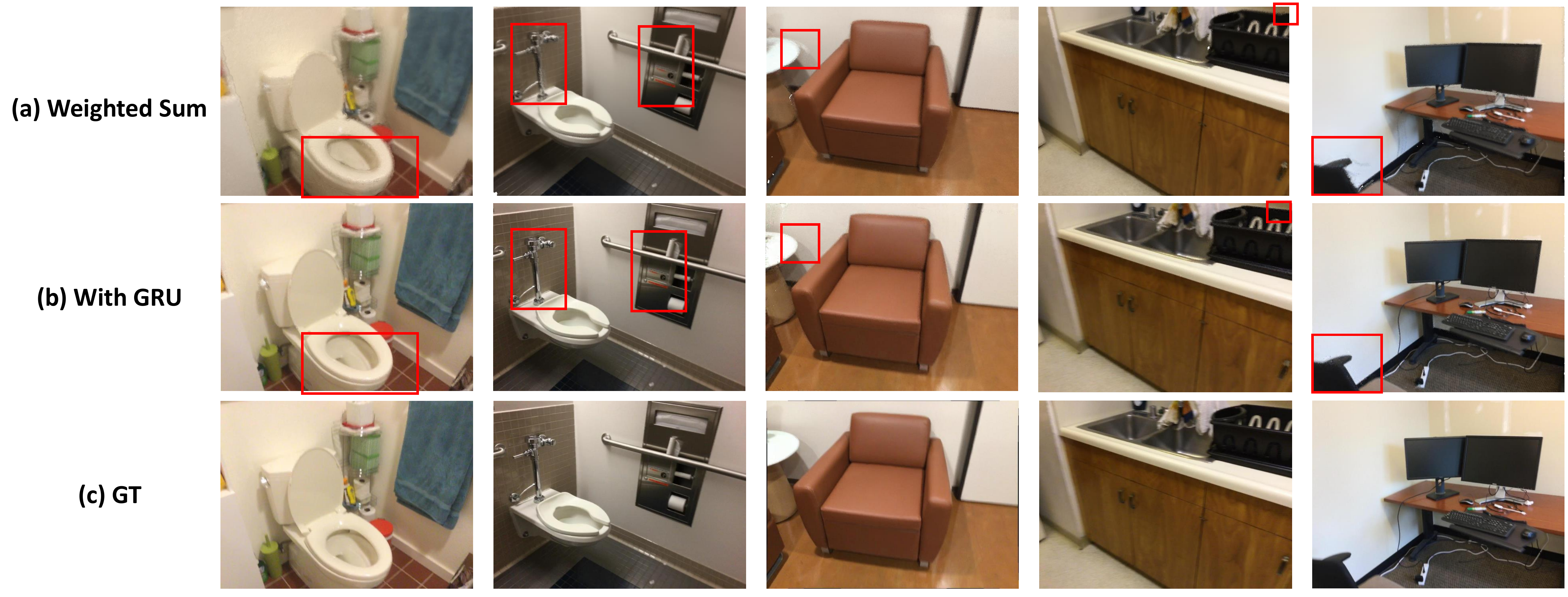}
	\end{minipage}
	\caption{Comparison of different fusion schemes on the per-scene optimization setting. The highlight areas are indicated by red rectangles. As can be seen in the figure, GRU can generate sharper and clearer details in novel view synthesis. GRU has the capability to adaptively update features based on high-level features, which makes the fusion process more robust. }
	\label{fig:gru}
\end{figure*}

\noindent\textbf{Fusion scheme.}
To investigate the effectiveness of the GRU fusion module, we have conducted an ablation study and shown quantitative results in the main paper. We recap the results which is shown in Table.~\ref{tab:ablation_surfel_fusion} and provide the qualitative results in Figure.~\ref{fig:gru}.

\begin{table}[t]
	\small
	\center
	\begin{center}
		\begin{tabular}{ccccc}
			\toprule
			Fusion Scheme  & & PSNR$\uparrow$ & SSIM$\uparrow$ &LPIPS $\downarrow$  \\ \midrule
			Weighted Sum  & No per-scene       & 23.09 & 0.833  & 0.353   \\ 
			GRU           & optimization  &  23.82 & 0.845  & 0.327  \\ 
			\midrule
			Weighted Sum  & Per-scene       & 28.54 & 0.884  & 0.293   \\ 
			GRU           & optimization  &\textbf{29.58}  & \textbf{0.919}  &0.215  \\ \bottomrule
		\end{tabular}
	\end{center}
	\caption{Ablation studies about fusion schemes in our  SurfelNeRF.}
	\label{tab:ablation_surfel_fusion}
\end{table}

\end{document}